# Empirical Methods for Compound Splitting

**Philipp Koehn**
Information Sciences Institute
Department of Computer Science
University of Southern California
`koehn@isi.edu`

**Kevin Knight**
Information Sciences Institute
Department of Computer Science
University of Southern California
`knight@isi.edu`

## Abstract

Compounded words are a challenge for NLP applications such as machine translation (MT). We introduce methods to learn splitting rules from monolingual and parallel corpora. We evaluate them against a gold standard and measure their impact on performance of statistical MT systems. Results show accuracy of 99.1% and performance gains for MT of 0.039 BLEU on a German-English noun phrase translation task.

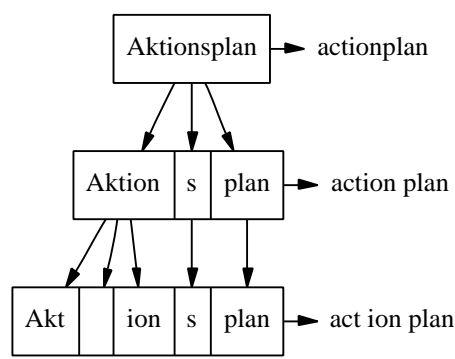

Figure 1: Splitting options for the German word *Aktionsplan*

## 1 Introduction

Compounding of words is common in a number of languages (German, Dutch, Finnish, Greek, etc.). Since words may be joined freely, this vastly increases the vocabulary size, leading to sparse data problems. This poses challenges for a number of NLP applications such as machine translation, speech recognition, text classification, information extraction, or information retrieval.

For machine translation, the splitting of an unknown compound into its parts enables the translation of the compound by the translation of its parts.

Take the word *Aktionsplan* in German (see Figure 1), which was created by joining the words *Aktion* and *Plan*. Breaking up this compound would assist the translation into English as *action plan*.

Compound splitting is a well defined computational linguistics task. One way to define the goal of compound splitting is to break up foreign words, so that a one-to-one correspondence to English can be established. Note that we are looking for a one-to-one correspondence to English content words: Say, the preferred translation of *Aktionsplan* is *plan for action*. The lack of correspondence for the English word *for* does not detract from the definition of the task: We would still like to break up the German compound into the two parts *Aktion* and *Plan*. The insertion of function words is not our concern.

Ultimately, the purpose of this work is to improve the quality of machine translation systems. For instance, phrase-based translation systems [Marcu and Wong, 2002] may recover more easily from splitting regimes that do not create a one-to-one translation correspondence. One splitting method may mistakenly break up the word *Aktionsplan* into the three words *Akt*, *Ion*, and *Plan*. But if we consistently break up the word *Aktion* into *Akt* and *Ion* in our training data, such a system will likely learn the translation of the word pair *Akt Ion* into the single English word *action*.

These considerations lead us to three different objectives and therefore three different evaluation metrics for the task of compound splitting:

- One-to-One correspondence
- Translation quality with a word-based translation system
- Translation quality with a phrase-based translation system

For the first objective, we compare the output of our methods to a manually created gold standard. For the second and third, we provide differently prepared training corpora to statistical machine translation systems.

## 2 Related Work

While the linguistic properties of compounds are widely studied [Langer, 1998], there has been only limited work on empirical methods to split up compounds for specific applications.

Brown [2002] proposes a approach guided by a parallel corpus. It is limited to breaking compounds into cognates and words found in a translation lexicon. This lexicon may also be acquired by training a statistical machine translation system. The methods leads to improved text coverage of an example based machine translation system, but no results on translation performance are reported.

Monz and de Rijke [2001] and Hedlund et al. [2001] successfully use lexicon based approaches to compound splitting for information retrieval. Compounds are broken into either the smallest or the biggest words that can be found in a given lexicon.

Larson et al. [2000] propose a data-driven method that combines compound splitting and word recombination for speech recognition. While it reduces the number of out-of-vocabulary words, it does not improve speech recognition accuracy.

Morphological analyzers such as Morphix [Finkler and Neumann, 1998] usually provide a variety of splitting options and leave it to the subsequent application to pick the best choice.

## 3 Splitting Options

Compounds are created by joining existing words together. Thus, to enumerate all possible splittings of a compound, we consider all splits into known words. Known words are words that exist in a training corpus, in our case the European parliament proceedings consisting of 20 million words of German [Koehn, 2002].

When joining words, filler letters may be inserted at the joint. These are called *Fugenelemente* in German. Recall the example of *Aktionsplan*, where the letter *s* was inserted between *Aktion* and *Plan*. Since there are no simple rules for when such letters may be inserted we allow them between any two words. As fillers we allow *s* and *es* when splitting German words, which covers almost all cases. Other transformations at joints include dropping of letters, such as when *Schweigen* and *Minute* are joined into *Schweigeminute*, dropping an *n*. A extensive study of such transformations is carried out by Langer [1998] for German.

To summarize: We try to cover the entire length of the compound with known words and fillers between words. An algorithm to break up words in such a manner could be implemented using dynamic programming, but since computational complexity is not a problem, we employ an exhaustive recursive search. To speed up word matching, we store the known words in a hash based on the first three letters. Also, we restrict known words to words of at least length three.

For the word *Aktionsplan*, we find the following splitting options:

- aktionsplan
- aktion–plan
- aktions–plan
- akt–ion–plan

We arrive at these splitting options, since all the parts – *aktionsplan, aktions, aktion, akt, ion,* and *plan* – have been observed as whole words in the training corpus.

These splitting options are the basis of our work. In the following we discuss methods that pick one of them as the correct splitting of the compound.

## 4 Frequency Based Metric

The more frequent a word occurs in a training corpus, the bigger the statistical basis to estimate translation probabilities, and the more likely the correct translation probability distribution is learned [Koehn and Knight, 2001]. This insight leads us to define a splitting metric based on word frequency.

Given the count of words in the corpus, we pick the split $S$ with the highest geometric mean of word frequencies of its parts $p_i$ ($n$ being the number of parts):

$$\text{argmax}_S \left( \prod_{p_i \in S} \text{count}(p_i) \right)^{\frac{1}{n}} \qquad (1)$$

Since this metric is purely defined in terms of German word frequencies, there is not necessarily a relationship between the selected option and correspondence to English words. If a compound occurs more frequently in the text than its parts, this metric would leave the compound unbroken – even if it is translated in parts into English.

In fact, this is the case for the example *Aktionsplan*. Again, the four options:

- aktionsplan(852) → 852
- aktion(960)–plan(710) → 825.6
- aktions(5)–plan(710) → 59.6
- akt(224)–ion(1)–plan(710) → 54.2

Behind each part, we indicated its frequency in parenthesis. On the right side is the geometric mean score of these frequencies. The score for the unbroken compound (852) is higher than the preferred choice (825.6).

On the other hand, a word that has a simple one-to-one correspondence to English may be broken into parts that bear little relation to its meaning. We can illustrate this on the example of *Freitag* (English: *Friday*), which is broken into *frei* (English: *free*) and *Tag* (English: *day*):

- frei(885)–tag(1864) → 1284.4
- freitag(556) → 556

## 5 Guidance from a Parallel Corpus

As stated earlier, one of our objectives is the splitting of compounds into parts that have one-to-one correspondence to English. One source of information about word correspondence is a parallel corpus: text in a foreign language, accompanied by translations into English. Usually, such a corpus is provided in form of sentence translation pairs.

Going through such a corpus, we can check for each splitting option if its parts have translations in the English translation of the sentence. In the case of *Aktionsplan* we would expect the words *action* and *plan* on the English side, but in case of *Freitag* we would not expect the words *free* and *day*. This would lead us to break up *Aktionsplan*, but not *Freitag*. See Figure 2 for illustration of this method.

This approach requires a translation lexicon. The easiest way to obtain a translation lexicon is to learn it from a parallel corpus. This can be done with the toolkit Giza [Al-Onaizan et al., 1999], which establishes word-alignments for the sentences in the two languages.

With this translation lexicon we can perform the method alluded to above: For each German word, we consider all splitting options. For each splitting option, we check if it has translations on the English side.

To deal with noise in the translation table, we demand that the translation probability of the English word given the German word be at least 0.01. We also allow each English word to be considered only once: If it is taken as evidence for correspondence to the first part of the compound, it is excluded as evidence for the other parts. If multiple options match the English, we select the one(s) with the most splits and use word frequencies as the ultimate tie-breaker.

**Second Translation Table**

While this method works well for the examples *Aktionsplan* and *Freitag*, it failed in our experiments for words such as *Grundrechte* (English: *basic rights*). This word should be broken into the two parts *Grund* and *Rechte*. However, *Grund* translates usually as *reason* or *foundation*. But here we are looking for a translation into the adjective *basic* or *fundamental*. Such a translation only occurs when *Grund* is used as the first part of a compound.

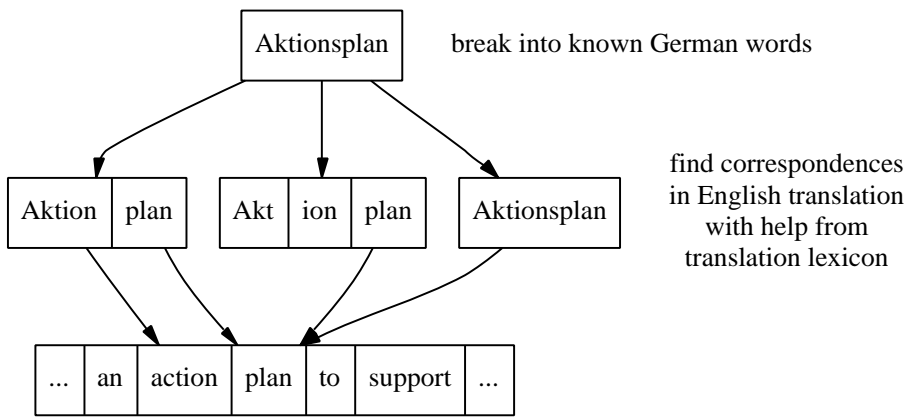

Figure 2: Acquisition of splitting knowledge from a parallel corpus: The split *Aktion–plan* is preferred since it has most coverage with the English (two words overlap)

To account for this, we build a second translation lexicon as follows: First, we break up German words in the parallel corpus with the frequency method. Then, we train a translation lexicon using Giza from the parallel corpus with split German and unchanged English.

Since in this corpus *Grund* is often broken off from a compound, we learn the translation table entry *Grund↔basic*. By joining the two translation lexicons, we can apply the same method, but this time we correctly split *Grundrechte*.

By splitting all the words on the German side of the parallel corpus, we acquire a vast amount of splitting knowledge (for our data, this covers 75,055 different words). This knowledge contains for instance, that *Grundrechte* was split up 213 times, and kept together 17 times.

When making splitting decisions for new texts, we follow the most frequent option based on the splitting knowledge. If the word has not been seen before, we use the frequency method as a back-off.

## 6 Limitation on Part-Of-Speech

A typical error of the method presented so far is that prefixes and suffixes are often split off. For instance, the word *folgenden* (English: *following*) is broken off into *folgen* (English: *consequences*) and *den* (English: *the*). While this is nonsensical, it is easy to explain: The word *the* is commonly found in English sentences, and therefore taken as evidence for the existence of a translation for *den*.

Another example for this is the word *Voraussetzung* (English: *condition*), which is split into *vor* and *aussetzung*. The word *vor* translates to many different prepositions, which frequently occur in English.

To exclude these mistakes, we use information about the parts-of-speech of words. We do not want to break up a compound into parts that are prepositions or determiners, but only content words: nouns, adverbs, adjectives, and verbs.

To accomplish this, we tag the German corpus with POS tags using the TnT tagger [Brants, 2000]. We then obtain statistics on the parts-of-speech of words in the corpus. This allows us to exclude words based on their POS as possible parts of compounds. We limit possible parts of compounds to words that occur most of the time as one of following POS: ADJA, ADJD, ADV, NN, NE, PTKNEG, VVFIN, VVIMP, VVINF, VVIZU, VVPP, VAFIN, VAIMP, VAINF, VAPP, VMFIN, VMINF, VMPP.

## 7 Evaluation

The training set for the experiments is a corpus of 650,000 noun phrases and prepositional phrases (NP/PP). For each German NP/PP, we have a English translation. This data was extracted from the Europarl corpus [Koehn, 2002], with the help of a German and English statistical parser. This limita-tion is purely for computational reasons, since we expect most compounds to be nouns. An evaluation of full sentences is expected to show similar results.

| Method | Correct | | Wrong | | | Metrics | | |
|---|---|---|---|---|---|---|---|---|
| | split | not | not | faulty | split | prec. | recall | acc. |
| raw | 0 | 3296 | 202 | 0 | 0 | - | 0.0% | 94.2% |
| eager | 148 | 2901 | 3 | 51 | 397 | 24.8% | 73.3% | 87.1% |
| frequency based | 175 | 3176 | 19 | 8 | 122 | 57.4% | 86.6% | 95.7% |
| using parallel | 180 | 3270 | 13 | 9 | 27 | 83.3% | 89.1% | 98.6% |
| **using parallel and POS** | 182 | 3287 | 18 | 2 | 10 | 93.8% | 90.1% | **99.1**% |

Table 1: Evaluation of the methods compared against a manual annotated gold standard of splits: Using knowledge from parallel corpus and part-of-speech information gives the best accuracy (99.1%).

We evaluate the performance of the described methods on a blind test set of 1000 NP/PPs, which contain 3498 words. Following good engineering practice, the methods have been developed with a different development test set. This restrains us from over-fitting to a specific test set.

### 7.1 One-to-one Correspondence

Recall that our first objective is to break up German words into parts that have a one-to-one translation correspondence to English words. To judge this, we manually annotated the test set with correct splits. Given this gold standard, we can evaluate the splits proposed by the methods.

The results of this evaluation are given in Table 1. The columns in this table mean:

**correct split:** words that should be split and were split correctly

**correct non:** words that should not be split and were not

**wrong not:** words that should be split but were not

**wrong faulty split:** words that should be split, were split, but wrongly (either too much or too little)

**wrong split:** words that should not be split, but were

**precision:** (correct split) / (correct split + wrong faulty split + wrong superfluous split)

**recall:** (correct split) / (correct split + wrong faulty split + wrong not split)

**accuracy:** (correct) / (correct + wrong)

To briefly review the methods:

**raw:** unprocessed data with no splits

**eager:** biggest split, i.e., the split into as many parts as possible. If multiple biggest splits are possible, the one with the highest frequency score is taken.

**frequency based:** split into most frequent words, as described in Section 4

**using parallel:** split guided by splitting knowledge from a parallel corpus, as described in Section 5

**using parallel and POS:** as previous, with an additional restriction on the POS of split parts, as described in Section 6

Since we developed our methods to improve on this metric, it comes as no surprise that the most sophisticated method that employs splitting knowledge from a parallel corpus and information about POS tags proves to be superior with 99.1% accuracy. Its main remaining source of error is the lack of training data. For instance, it fails on more obscure words such as *Passagier–aufkommen* (English: *passenger volume*), where even some of the parts have not been seen in the training corpus.

### 7.2 Translation Quality with Word Based Machine Translation

The immediate purpose of our work is to improve the performance of statistical machine translation systems. Hence, we use the splitting methods to prepare training and testing data to optimize the performance of such systems.

| Method | BLEU |
|---|---|
| raw | 0.291 |
| eager | 0.222 |
| **frequency based** | **0.317** |
| using parallel | 0.294 |
| using parallel and POS | 0.306 |

Table 2: Evaluation of the methods with a word based statistical machine translation system (IBM Model 4). Frequency based splitting is best, the methods using splitting knowledge from a parallel corpus also improve over unsplit (raw) data.

| Method | BLEU |
|---|---|
| raw | 0.305 |
| **eager** | **0.344** |
| **frequency based** | **0.342** |
| using parallel | 0.330 |
| using parallel and POS | 0.326 |

Table 3: Evaluation of the methods with a phrase based statistical machine translation system. The ability to group split words into phrases overcomes the many mistakes of maximal (eager) splitting of words and outperforms the more accurate methods.

First, we measured the impact on a word based statistical machine translation system, the widely studied IBM Model 4 [Brown et al., 1990], for which training tools [Al-Onaizan et al., 1999] and decoders [Germann et al., 2001] are freely available. We trained the system on the 650,000 NP/PPs with the Giza toolkit, and evaluated the translation quality on the same 1000 NP/PP test set as in the previous section. Training and testing data was split consistently in the same way. The translation accuracy is measured against reference translations using the BLEU score [Papineni et al., 2002]. Table 2 displays the results.

Somewhat surprisingly, the frequency based method leads to better translation quality than the more accurate methods that take advantage from knowledge from the parallel corpus. One reason for this is that the system recovers more easily from words that are split too much than from words that are not split up sufficiently. Of course, this has limitations: Eager splitting into as many parts as possible fares abysmally.

### 7.3 Translation Quality with Phrase Based Machine Translation

Compound words violate the bias for one-to-one word correspondences of word based SMT systems. This is one of the motivations for phrase based systems that translate groups of words. One of such systems is the joint model proposed by Marcu and Wong [2002]. We trained this system with the different flavors of our training data, and evaluated the performance as before. Table 3 shows the results.

Here, the eager splitting method that performed so poorly with the word based SMT system comes out ahead. The task of deciding the granularity of good splits is deferred to the phrase based SMT system, which uses a statistical method to group phrases and rejoin split words. This turns out to be even slightly better than the frequency based method.

## 8 Conclusion

We introduced various methods to split compound words into parts. Our experimental results demonstrate that what constitutes the optimal splitting depends on the intended application. While one of our method reached 99.1% accuracy compared against a gold standard of one-to-one correspondences to English, other methods show superior results in the context of statistical machine translation. For this application, we could dramatically improve the translation quality by up to 0.039 points as measured by the BLEU score.

The words resulting from compound splitting could also be marked as such, and not just treated as regular words, as they are now. Future machine translation models that are sensitive to such linguistic clues might benefit even more.